# Decomposing predictability: Semantic feature overlap between words and the dynamics of reading for meaning

Markus J. Hofmann[1*], Mareike A. Kleemann[1], André Roelke[1],

Christian Vorstius[1], & Ralph Radach[1]

[1]General and Biological Psychology, University of Wuppertal, Max-Horkheimer-Str. 20,

42119 Wuppertal, Germany

*correspondence to mhofmann@uni-wuppertal.de

Word Count of the main text: 2992

Declarations of interest: none

**Abstract**

The present study uses a computational approach to examine the role of semantic constraints in normal reading. This methodology avoids confounds inherent in conventional measures of predictability, allowing for theoretically deeper accounts of semantic processing. We start from a definition of associations between words based on the significant log likelihood that two words co-occur frequently together in the sentences of a large text corpus. Direct associations between stimulus words were controlled, and semantic feature overlap between *prime* and **target** words was manipulated by their common associates. The stimuli consisted of sentences of the form pronoun, *verb*, article, *adjective* and **noun**, followed by a series of closed class words, e. g. "She *rides* the *grey* **elephant** on one of her many exploratory voyages". The results showed that *verb*-**noun** overlap reduces single and first fixation durations of the target noun and *adjective*-**noun** overlap reduces go-past durations. A dynamic spreading of activation account suggests that associates of the *prime* words take some time to become activated: The *verb* can act on the **target noun's** early eye-movement measures presented three words later, while the *adjective* is presented immediately prior to the **target**, which induces sentence re-examination after a difficult adjective-noun semantic integration.

1. Introduction

McDonald and Shillcock (2003a) introduced statistical text corpus information to the field of eye movement research on reading, using the transitional probability of a word to occur, given the previous word. In this case, word contiguity is taken to define an associative relation of a prime word occurring immediately prior to a target word (e.g. Hofmann et al., 2018). In contrast, semantic feature overlap of two words can be computed from the number of their common associates (CA; e.g. Evert, 2005; Roelke et al., 2018). In that case, each common associate can be defined as a semantic feature. The present study reports the first work extending this approach to the realm of natural reading. We presented skilled readers with pronoun-verb-article-adjective-noun sentences, while experimentally manipulating the number of CA of verb and adjective to the target noun.

The "classic" approach to contextual constraints in sentence processing relies on empirical predictability obtained from incremental close tasks (e.g., Taylor, 1953). Since Ehrlich and Rayner (1981), cloze completion probability (CCP) has been shown to influence differential eye movement parameters, though there was considerable variability between studies on whether earlier or later eye movement parameters are affected (Staub, 2015). When aiming to investigate which word is constraining certain other words, however, this empirical approach to sentence semantics is limited. Importantly, CCP is an "all-in" variable, confounding syntactic, contiguity-based (associative) and semantic constraints, as discussed in some detail by Staub (2015).

Computational approaches allow studying more specific types of inter-word relations. They take a text corpus for training, from which the computational models consolidate a long-term memory structure, generating predictions tested at retrieval (Hofmann et al., 2018). Using transitional probabilities, McDonald and Schillcock (2003a) paired verbs with likely and less likely nouns and showed significant effects on single-fixation duration (SFD) and first fixation duration (FFD). McDonald and Shillcock (2003b) suggest that transitional probability reflects a low-level process, while it does not capture high-level conceptual knowledge that is probably better reflected by CCP. Stronger constraints can be obtained by considering more than one prime word: In an n-gram model, the $n^{th}$ word is predicted by the preceding n-1 words. Smith and Levy's (2013) 3-gram model was able to predict gaze duration (GD), suggesting that multiple contiguities may influence later components of eye movement control.

While word contiguity measures are relatively straightforward to define syntactic and associative relations, the challenge arises how to compute higher-level semantic structure. A simple approach to semantic feature overlap is provided by the number of contiguous words that two words share to define e.g. synonymy (Rapp, 2002). Based on a such second-order contiguity, Landauer and Dumais (1997) computed latent semantic dimensions determining higher-order contiguities in documents. Latent semantic analysis (LSA) predicted sentence comprehension and performed similar to children in synonym judgment. Pynte, New and Kennedy (2008) demonstrated LSA-based SFD- and GD-effects using a set of newspaper texts. Wang, Pomplun, Chen, Ko and Rayner (2010) found that transitional probability affects early FFD and GD effects, while LSA was related to later total viewing duration (TVD). LSA has been challenged by a Bayesian account: Griffiths, Steyvers and Tennenbaum (2007) showed that topics models can better predict semantic priming and disambiguation effects in GD and TVD data (cf. Sereno, Pacht & Rayner, 1992).

Recurrent neural network models (RNNs) are a further possibility to computationally define syntactic and semantic word features: Elman (1990) trained a set of recurrent hidden units to learn statements such as "A robin is a bird". RNNs learn to gradually differentiate between syntactic classes such as verbs and nouns, non-living and living objects, and between mammals and fishes. Frank (2009) found that an RNN provides significantly larger correlations with GD compared to a surprisal measure of the grammatical category of the word. An RNN, a topics model and a 5-gram model together explained not only about half of the variance of CCP, but they also performed significantly better than CCP in predicting SFD data (Hofmann, Biemann, & Remus, 2017).

Most computational studies compute contiguity-based (associative), semantic and/or syntactic constraints for a given set of reading materials, and then examine fixation parameters using a regression-based approach. This may lead to serious confounds (see Rayner, Pollatsek, Drieghe, Slattery, & Reichle, 2007, for a discussion) that are circumvented in the experimental approach. We use a single computational approach that allows to disentangle associative contiguity and semantic feature overlap, while keeping syntactic structure constant. Word contiguity was defined by two words co-occurring significantly more often together in sentences than expected from single-occurrence frequencies (Dunning, 1993). Hofmann, Kuchinke, Biemann, Tamm and Jacobs (2011) relied on this log likelihood test to define between-word associations: If the words do not co-occur significantly more often together, they provide an association strength (AS) of 0. If they significantly co-occur in sentences, they are associated and AS is defined as the log-transformed $\chi^2$-value. Based on this simple approach, semantic feature overlap can be computed by counting the number of CA (Roelke, Franke, Radach, Biemann, & Hofmann, 2018; cf. Evert, 2005). The number of associated stimuli in a recognition memory task can predict false and veridical memory effects in behavioral (e.g. Hofmann et al., 2011; Hofmann & Jacobs, 2014; cf. Roediger & McDermott, 1995), fMRI (Kuchinke, Fritzemeier, Hofmann, & Jacobs, 2013) and ERP data (Stuellein, Radach, Jacobs, & Hofmann, 2016). AS between two nouns can predict association ratings (Hofmann et al., 2018) and neural activation in the left inferior frontal gyrus (Hofmann & Jacobs, 2014). Moreover, AS and the number of CA can predict associative and semantic priming effects during lexical decision in adults and children (Franke, Roelke, Radach & Hofmann, 2017; Roelke et al., 2018; Figure 1).

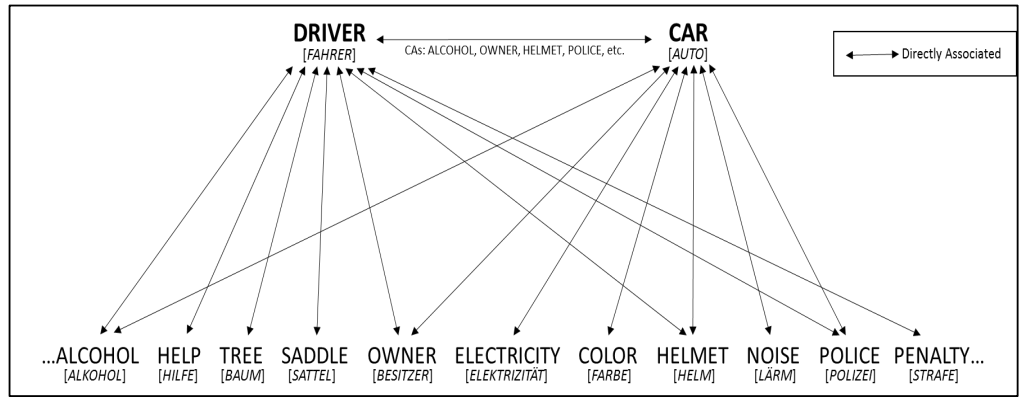

Fig. 1. A simple approach to disentangle contiguity and semantic feature overlap of two words. Driver and car often co-occur in the same sentence, and therefore are associated, but they also contain many common associates, e.g. alcohol, owner and helmet (cf. e.g. Roelke et al., 2018).

Roelke et al. (2018) showed that AS can predict associative priming at a short (200 ms) and at a long (1000 ms) stimulus-onset asynchrony (SOA), but the number of CA elicits semantic priming only at the long SOA. This RT pattern is similarly observed in the classic priming literature (Ferrand & New, 2003; Hutchison, 2003; Lucas, 2000). Differential processes may contribute to long-SOA priming: Facilitation is observed when a semantic expectancy is met, while strong semantic competitors can also lead to inhibition (Neely, 1977). Balota, Black and Cheney (1992) ask whether lexical decisions are optimal for investigating semantic priming, because they are influenced by postlexical checking.

In the present reading experiment, we expected stronger semantic priming at a short than at long SOA (Roelke et al., 2018). In their early sentence reading study, Carrol and Slowiaczek (1986) found that category primes, sharing many features with high-typicality target exemplars, induce a large mean fixation duration facilitation (41 ms) at a short SOA, in which the mean fixation onsets of the prime and target differed by 488 ms. At a much longer SOA of 1247 ms, they observed a 10 ms facilitation. However, it is not clear what the definition of their mean fixation duration measure is and whether it might confound differential types of fixation cases (Inhoff & Radach, 1998). The present study will address early and later lexical processing separately by analyzing single fixation duration (SFD), first fixation duration (FFD) and first-pass gaze duration GD (see Radach & Kennedy, 2004, for definitions). As total viewing time (TVD) included target re-reading after visiting non-controlled words to the right, this parameter is only reported but not discussed. Rather, we focused on go-past duration (GPD; e.g. Schotter, 2013), a measure that includes leftward regressions to account for late repair and integration effort. We were careful to present at last three closed-class words after the target to minimize any potential parafovea-on-fovea effects (e.g. Radach, Inhoff, Glover, & Vorstius, 2013). At the end of our sentences, we added a number of further open- and closed-class words to provide meaningful stimuli.

## 2. Method

*2.1 Participants*

32 German native speakers with normal or corrected-to normal vision and without language disorders participated in the study for cash or course credits. Two participants were excluded, because they deviated more than 2 SD ($SD = 3.76$) from the mean error score ($M = 6.38$; range = 0 to 17) in the comprehension test. The remaining 30 subjects had a mean age of 23.60 years ($SD = 5.86$, range = 19 to 44, 21 female).

*2.2 Materials*

Stimuli consisted of 160 German pronoun-verb-article-adjective-noun sentences, continued with three closed-class words (articles, prepositions, conjunctions or pronouns) and 1-6 additional words. 40 filler sentences contained no open-class words from the experimental stimuli and did not follow any syntactic construction rule. Sentences consisted of 69-72 characters and 9-14 words. Verb-noun and adjective-noun semantic feature overlap was manipulated by the number of CA (Hofmann et al., 2018). All computations were based on the lemmas accumulated in the German corpus of the Leipzig Wortschatz Project[1] (70 million sentences, 1.1 billion words; Goldhahn, Eckart & Quasthoff, 2012). We used the 1000 words with the largest AS (Dunning, 1993; Hofmann et al., 2011) and counted the number of CA of each prime-target pair. To constrain the CA to words relatively diagnostic for a particular meaning, we excluded the 100 most frequent words (Griffiths et al., 2007; Hofmann et al., 2018). With the second experimental factor of verb vs. adjective, this resulted in four experimental conditions each containing 40 sentences (Table 1).

---

[1] http://www.corpora.uni-leipzig.de/en?corpusId=deu_newscrawl-public_2018

Table 1. Example sentences for the experimental conditions

| Prime: | CA with target | | Example |
| --- | --- | --- | --- |
| | Verb | Adjective | |
| SOA: | Long | Short | |
| HH | High | High | Sie *reitet* den *grauen* **Elefanten** auf einer ihrer vielen Forschungsreisen. (She *rides* the *grey* **elephant** on one of her many exploratory voyages.) |
| HL | High | Low | Er *zeigt* das *amtliche* **Muster** seinen in einem Büroraum wartenden Kollegen. (He *shows* the *official* **sample** to his colleagues waiting in an office room.) |
| LH | Low | High | Sie *testet* den *flinken* **Frosch** mit einer von ihr entwickelten Messmethode. (She *tests* the *swift* **frog** with a measuring method developed by her). |
| LL | Low | Low | Er *erwarb* das *klapprige* **Gefährt** mit einem seiner ungedeckten Schecks. (He *acquired* the *shaky* **vehicle** with one of his uncovered checks.) |

*Note:* CA – number of common associates of *prime* and **target** words. High: CA > 60; Low: CA < 15 CA.

We controlled Leipzig word frequency classes relating the frequency of each word to the frequency of the most frequent German word: "der" [the] is $2^{class}$ more frequent than the given word (Goldhahn et al., 2012). Length, frequency, and number of orthographic neighbors (ON) of the prime and target words, prime-target AS, AS and CA between the primes, as well as length and frequency of the closed-class words after the target were experimentally controlled (all $F$s < 1, see Table 2 and Appendix A). Half of the experimental trials in each stimulus category included a comma after the target to obtain syntactic variety.

Table 2. Mean values (*SD* in parentheses) of manipulated and controlled variables

|  | HH | HL | LH | LL |
|---|---|---|---|---|
| CA Verb-noun | 78.93 (15.82) | 78.28 (18.47) | 10.88 (3.33) | 10.68 (3.05) |
| CA Adjective-noun | 86.05 (23.57) | 10.38 (3.61) | 85.08 (25.72) | 11.18 (3.86) |
| AS Verb-noun | 1.27 (0.32) | 1.26 (0.32) | 1.19 (0.31) | 1.23 (0.26) |
| AS Adjective-noun | 1.24 (0.36) | 1.19 (0.27) | 1.26 (0.33) | 1.16 (0.26) |
| Target noun | | | | |
| Length | 6.10 (1.43) | 6.15 (1.33) | 6.23 (1.21) | 5.98 (1.46) |
| Frequency | 11.38 (1.69) | 10.90 (1.84) | 11.20 (1.77) | 11.53 (2.72) |
| ON | 1.88 (3.20) | 1.68 (1.95) | 1.30 (1.96) | 1.53 (2.16) |
| Verb | | | | |
| Length | 7.03 (0.92) | 6.88 (1.07) | 7.18 (0.90) | 7.03 (0.86) |
| Frequency | 12.33 (1.87) | 12.65 (2.39) | 12.85 (3.85) | 12.60 (3.69) |
| ON | 2.58 (2.95) | 2.33 (2.76) | 1.85 (2.15) | 2.20 (2.40) |
| Adjective | | | | |
| Length | 6.45 (1.11) | 6.53 (1.18) | 6.25 (1.41) | 6.38 (1.37) |
| Frequency | 13.65 (1.31) | 13.25 (4.06) | 12.98 (2.03) | 13.23 (3.13) |
| ON | 0.55 (0.81) | 0.73 (1.11) | 0.75 (1.32) | 0.90 (1.17) |

*Note*: AS = association strength; ON = number of orthographic neighbors.

*2.3 Procedure*

Participants were instructed to read silently at their normal pace, so that they were able to respond to comprehension questions. Eye movements were recorded by an EyeLink1000® (2000 Hz, SR Research, Toronto, Canada). Participants used a chin- and forehead-rest to minimize head movements. Three-point calibration was performed at the beginning of the experiment, after every block and after each comprehension question. The experiment started with 12 practice trials. Each trial started with a fixation point presented one letter to the left of the beginning of the first word, simultaneously serving as drift check. Deviations greater 0.33° triggered an additional calibration. Sentences were displayed as single lines in black font (Courier New, 18 pt) on a light-grey background, vertically centered on a 24-inch flat panel monitor (1680 x 1050 pixel, 120 Hz; viewing distance: 68.75 cm). A letter corresponded to a visual angle of 0.33°. 1000 ms blank screens were presented after participants initiated the next trial by button press, which were followed by comprehension questions after each practice trial and after a randomly selected third of the main experimental trials (67 questions) – they were answered orally. The 200 sentences were pseudorandomized in two lists with no more than two sentences of the same experimental condition to appear consecutively. We split these lists into two blocks, making sure that the first and second block had approximately the same number of sentences of each category, also balancing list and block order across participants.

*2.4 Analyses*

Right-eye fixations on the critical target word were analyzed if both primes were fixated before. We removed fixations <70 ms and >800 ms for SFD and FFD, >1000 ms for GD, and >1500 ms for GPD and TVD. Inferential statistics were based on linear mixed models (LMMs) with maximum likelihood estimation (*lme4* and *lmerTest* packages in R). Fixed effects were the number of CA of verb and adjective to the target word (low vs. high) and their interaction, using successive differences coding (-0.5 vs. 0.5; contr.sdif, *MASS* package). LMMs always started with a maximum random structure including random slopes for both effects (Barr, Levy, Scheeperes, & Tily, 2013). Then we simplified LMMs by removing random slopes for interactions and main effects, which either led to singular matrices or failure to converge (cf. Baayen, 2007). The final models contained random item and subject intercepts (Bates, Kliegl, Vasishth, & Baayen, 2015). We removed trials in which residuals deviated more than 2.5 SD from mean (see Table 3, for the trials remaining for analyses). Kolmogorov-Smirnov tests for all final models indicated no significant deviance from normality (all *P*s > 0.05), except for the GD and TVD analyses (*Ps* = 0.007). Eye movement data were log-transformed for inferential statistics, but Figures and Tables report non-transformed values.

Table 3. Means (*SE*) of the target noun for the different eye movement parameters.

|     | HH | HL | LH | LL |
|-----|-----|-----|-----|-----|
| SFD | 235 (3) | 233 (3) | 240 (3) | 244 (3) |
| FFD | 234 (3) | 232 (3) | 240 (3) | 244 (3) |
| GD  | 262 (4) | 257 (4) | 265 (4) | 280 (5) |
| TVD | 419 (7) | 374 (7) | 355 (8) | 358 (9) |
| GPD | 601 (11) | 633 (15) | 637 (13) | 710 (14) |

*Note*. SFD = Single fixation duration; FFD = First fixation duration; GD = Gaze duration; TVD = Total viewing duration GPD = Go-past duration.

Table 4. Results of the LMM analyses (* P < 0.05)

| | N rows | Intercept | | CA Verb | | | CA Adjective | | | CA Verb * CA Adjective | | | Random Intercept $\sigma^2$ | | Residual $\sigma^2$ |
|---|---|---|---|---|---|---|---|---|---|---|---|---|---|---|---|
| | | B | SE | B | SE | T | B | SE | T | B | SE | T | Item | Subject | |
| SFD | 3179 | 5.41 | 0.03 | -0.03 | 0.02 | **-2.12** * | 0 | 0.02 | 0.08 | 0.03 | 0.03 | 1.04 | 0.01 | 0.03 | 0.07 |
| FFD | 3702 | 5.4 | 0.03 | -0.04 | 0.01 | **-2.54** * | 0 | 0.01 | -0.29 | 0.02 | 0.03 | 0.81 | 0.01 | 0.03 | 0.07 |
| GD | 3716 | 5.48 | 0.03 | -0.04 | 0.02 | -1.92 | -0.02 | 0.02 | -0.7 | 0.07 | 0.05 | 1.6 | 0.02 | 0.03 | 0.10 |
| TVD | 3722 | 5.77 | 0.05 | -0.07 | 0.03 | -1.95 | -0.04 | 0.03 | -1.2 | 0.11 | 0.07 | 1.61 | 0.04 | 0.06 | 0.17 |
| GPD | 1134 | 6.24 | 0.03 | -0.05 | 0.03 | -1.67 | -0.06 | 0.03 | **-1.98** * | 0.05 | 0.07 | 0.83 | 0.02 | 0.03 | 0.12 |

### 3. Results and Discussion

We examined differential reading time measures in target nouns, while experimentally manipulating the feature overlap with the preceding verb and adjective in pronoun-verb-article-adjective-noun sentences. A simple computational approach was used to manipulate semantic feature overlap using the number of CA (e.g. Hofmann et al., 2018), while controlling for AS between all open-class words and CA between the primes (Table 2; Appendix A). An average time of 1152 ms ($SE = 8$) passed between the verb prime and target noun fixation onsets in the long-SOA conditions. We found early priming effects of verb-noun semantic overlap in the SFD and FFD (Table 4). 7 and 8 ms facilitations for high verb-noun overlap was observed in SFD and FFD data, respectively ($SE$s = 3). The fixation onsets of the adjective and the noun differed by a $M = 515$ ms ($SE = 6$) in the short SOA conditions. Here we found an effect of adjective-noun semantic feature overlap in the GPD. A high overlap resulted in an average facilitation of 47 ms ($SE = 15$).

From the perspective of the semantic priming literature in visual word recognition, our results are straightforward: Long-SOA semantic priming typically elicits smaller facilitation than short-SOA priming (Lucas, 2000). In contrast to Roelke et al.'s (2018) primed lexical decision study, we now found long-SOA semantic priming effects during sentence reading. The missing long-SOA lexical decision effects may be explained either by a postlexical checking mechanism (Balota et al., 1992) or by a strong emphasis on prime processing inhibiting target processing (Plaut & Booth, 2000). During natural reading, in contrast, spreading activation is directed to subsequently presented stimulus words.

As opposed to the present study, RTs in non-natural-reading tasks do not allow to disentangle early and late processes during the dynamic processing of the target word. We observed verb-noun (long-SOA) priming in SFD and FFD data, reflecting an early stage of processing, while (short-SOA) adjective-noun priming elicited a late effect in the GPD. This result pattern can be explained by the time available for the spreading of semantic activation: When a prime is presented, its semantic features become active (Hofmann et al., 2011, Fig. 4; Radach & Hofmann, 2016, Fig. 2; McClelland & Rumelhart, 1981), but they need time to become sufficiently active for influencing the semantic features of the target. The stronger the activation of the semantic features of the prime, the more immediate will be the interaction with the target's semantic features. With a long SOA, the features of the verb prime have sufficient time to become active, leading to early SFD and FFD effects. At a short SOA, in contrast, the adjective's semantic features do not have enough time to become sufficiently active to elicit an early effect. After first-pass reading has been finished, however, a sufficient period of time has elapsed for the CA of the adjective and the noun to become active: In the high-CA condition, the adjective and the noun can be semantically integrated with ease, but in the low-CA condition semantic integration is more likely to fail. When assuming a semantic layer feeding activation to an orthographic layer (e.g. Hofmann & Jacobs, 2014; McNamara, 2005, p. 41), the preceding sentence context may gain saliency and is therefore more likely to be re-examined (Reilly & Radach, 2006; Snell, van Leipsig, Grainger & Meeter, 2018). This is reflected in the greater GPD durations for low adjective-noun overlap. Traxler, Foss, Seely, Kaup, & Morris (2000, experiment 1) reported a similar pattern of results: They presented sentences such as "The lumberjack *carried* the **axe** early in the morning". When replacing the verb by schema-inconsistent verbs like "chopped", they observed slower TVD, but no effects of FFD or GD – a relatively short prime-target SOA produced effects in a late eye movement measure only.

The 'all-in' variable of empirical predictability confounds syntactic, contiguity-based (associative) and semantic effects (e.g. Staub, 2015). Here, we constrained the functional locus to semantic priming, therefore addressing the semantic integration of particular word pairs. We believe that manipulating and controlling for differential computational parameters will be essential for obtaining a consistent and deep explanation for semantic effects in natural reading (cf. Reichle, Rayner, & Pollatsek, 2003, p. 450). This deeper theoretical knowledge could be considered in the development of future computational models of eye-movement control (cf. Engbert, Nuthmann, Richter, & Kliegl, 2005).

### 4. Acknowledgments

MJH, MK and RR planned and designed the study. MK and AR selected the stimuli. MJH, MK and CV performed the analyses. All authors wrote the paper (major writing: MJH and MK; minor writing: AR, CV and RR). We like to thank Albrecht Inhoff and Chris Biemann for inspiration while designing the study, which was partly supported by a grant to MJH and RR (DFG-Gz: HO5139/2-2; RA1603/4-2).

# 5. References

Balota, D. A., Black, S. R., & Cheney, M. (1992). Automatic and Attentional Priming in Young and Older Adults: Reevaluation of the Two-Process Model. *Journal of Experimental Psychology: Human Perception and Performance*, *18*(2), 485–502.

Barr, D. J., Levy, R., Scheepers, C., & Tily, H. J. (2013). Random effects structure for confirmatory hypothesis testing: Keep it maximal. *Journal of Memory and Language*, *68*(3), 255–278.

Bates, D., Kliegl, R., Vasishth, S., & Baayen, H. (2015). Parsimonious Mixed Models. *ArXiv:1506.04967*, (2000), 1–21.

Carroll, P., & Slowiaczek, M. L. (1986). Constraints on semantic priming in reading: A fixation time analysis. *Memory & Cognition*, *14*(6), 509–522.

Dunning, T. (1993). Accurate methods for the statistics of surprise and coincidence. *Computational Linguistics*, *19*, 61–74.

Ehrlich, S. F., & Rayner, K. (1981). Contextual effects on word perception and eye movements during reading. *Journal of Verbal Learning and Verbal Behavior*, *20*(6), 641–655.

Elman, J. (1990). Finding Structure in Time. *Cognitive Science*, *14*, 179–211.

Engbert, R., Nuthmann, A., Richter, E. M., & Kliegl, R. (2005). SWIFT: a dynamical model of saccade generation during reading. *Psychological Review*, *112*(4), 777–813.

Evert, S. (2005). *The Statistics of Word Cooccurrences Word Pairs and Collocations*. Universität Stuttgart, Stuttgart.

Ferrand, L., & New, B. (2003). Semantic and associative priming in the mental lexicon. In P. Bonin (Ed.), *Mental lexicon: "Some words to talk about words"* (pp. 25–43). New York: Nova science publishers.

Frank, S. (2009). Surprisal-based comparison between a symbolic and a connectionist model of sentence processing. In *Proceedings of the annual meeting of the Cognitive Science Society* (pp. 1139–1144).

Franke, N., Roelke, A., Radach, R. R., & Hofmann, M. J. (2017). After braking comes hasting: reversed effects of indirect associations in 2nd and 4th graders. In *Proceedings of the Cognitive Science Society* (pp. 2025–2030). London, UK.

Goldhahn, D., Eckart, T., & Quasthoff, U. (2012). Building large monolingual dictionaries at the leipzig corpora collection: From 100 to 200 languages. *Proceedings of the 8th International Conference on Language Resources and Evaluation*, 759–765.

Griffiths, T. L., Steyvers, M., & Tenenbaum, J. B. (2007). Topics in semantic representation. *Psychological Review*, *114*(2), 211–244.

Hofmann, M. J., Biemann, C., & Remus, S. (2017). Benchmarking n-grams, topic models and recurrent neural networks by cloze completions, EEGs and eye movements. In B. Sharp, F. Sedes, & W. Lubaszewsk (Eds.), *Cognitive Approach to Natural Language Processing* (pp. 197–215). London, UK: ISTE Press Ltd, Elsevier.

Hofmann, M. J., Biemann, C., Westbury, C. F., Murusidze, M., Conrad, M., & Jacobs, A. M. (2018). Simple co-occurrence statistics reproducibly predict association ratings. *Cognitive Science*, *42*, 2287–2312.

Hofmann, M. J., & Jacobs, A. M. (2014). Interactive activation and competition models and semantic context: From behavioral to brain data. *Neuroscience and Biobehavioral Reviews*, *46*, 85–104.

Hofmann, M. J., Kuchinke, L., Biemann, C., Tamm, S., & Jacobs, A. M. (2011). Remembering words in context as predicted by an associative read-out model. *Frontiers in Psychology*, *2*(252), 1–11.

Hutchison, K. A. (2003). Is semantic priming due to association strength or feature overlap ? A micro analytic review. *Psychonomic Bulletin & Review*, *10*(4), 785–813.

Inhoff, A. W., & Radach, R. (1998). Definition and computation of oculomotor measures in the study of cognitive processes. In G. Underwood (Ed.), *Eye Guidance in Reading and Scene Perception* (pp. 29–53). Oxford, England: Elsevier Science.

Kuchinke, L., Fritzemeier, S., Hofmann, M. J., & Jacobs, A. M. (2013). Neural correlates of episodic memory: Associative memory and confidence drive hippocampus activations. *Behavioural Brain Research*, *254*, 92–101.

Landauer, T. K., & Dumais, S. T. (1997). A solution to Platos problem: The latent semantic analysis theory of acquisition, induction and representation of knowledge. *Psychological Review*, *104*(2), 211–240.

Lucas, M. (2000). Semantic priming without association: a meta-analytic review. *Psychonomic Bulletin & Review*, *7*(4), 618–630.

McClelland, J. L., & Rumelhart, D. E. (1981). An interactive activation model of context effects in letter perception: Part 1. An account of basic findings. *Psychological Review*, *88*(5), 375–407.

McDonald, S. A., & Shillcock, R. C. (2003a). Eye movements reveal the on-line computation of lexical probabilities during reading. *Psychological Science*, *14*(6), 648–652.

McDonald, S. A., & Shillcock, R. C. (2003b). Low-level predictive inference in reading: The influence of transitional probabilities on eye movements. *Vision Research*, *43*(16), 1735–1751.

McNamara, T. P. (2005). *Semantic priming: Perspectives from memory and word recognition*. New York: Taylor & Francis Group.

Neely, J. H. (1977). Semantic priming and retrieval from lexical memory: Roles of inhibitionless spreading activation and limited-capacity attention. *Journal of Experimental Psychology: General*, *106*(3), 226–254.

Plaut, D. C., & Booth, J. R. (2000). Individual and developmental differences in semantic priming: Empirical and computational support for a single-mechanism account of lexical processing. *Psychological Review*, *107*(4), 786–823.

Pynte, J., New, B., & Kennedy, A. (2008). On-line contextual influences during reading normal text: A multiple-regression analysis. *Vision Research*, *48*(21), 2172–2183.

Radach, R., & Hofmann, M. J. (2016). Graphematische Verarbeitung beim Lesen von Wörtern. In U. Domahs & B. Primus (Eds.), *Laut, Gebärde, Buchstabe (Handbuch Sprachwissen, Band 2)* (pp. 455–473). Berlin: De Gruyter Mouton.

Radach, R., Inhoff, A. W., Glover, L., & Vorstius, C. (2013). Contextual constraint and N + 2 preview effects in reading. *Quarterly Journal of Experimental Psychology*, *66*(3), 619–633.

Radach, R., & Kennedy, A. (2004). European Journal of Cognitive Theoretical perspectives on eye movements in reading: Past controversies , current issues , and an agenda for future research. *European Journal of Cognitive Psychology*, *16*(1/2), 3–26.

Rapp, R. (2002). The Computation of Word Associations: Comparing Syntagmatic and Paradigmatic Approaches. In *Proceedings of the 19th international conference on Computational linguistics - Volume 1 (COLING)* (pp. 1–7).

Reichle, E. D., Rayner, K., & Pollatsek, A. (2003). The E-Z reader model of eye-movement control in reading: comparisons to other models. *The Behavioral and Brain Sciences*, *26*(4), 445–476.

Reilly, R. G., & Radach, R. (2006). Some empirical tests of an interactive activation model of eye movement control in reading. *Cognitive Systems Research*, *7*, 34–55.

Roediger, H. L., & McDermott, K. B. (1995). Creating false memories: Remembering words not presented in lists. *Journal of Experimental Psychology: Learning, Memory, and Cognition*, *21*(4), 803–814.

Roelke, A., Franke, N., Biemann, C., Radach, R., Jacobs, A. M., & Hofmann, M. J. (2018). A novel co-occurrence-based approach to predict pure associative and semantic priming. *Psychonomic Bulletin and Review*, *25*(4), 1488–1493.

Schotter, E. R. (2013). Synonyms provide semantic preview benefit in English. *Journal of Memory and Language*, *69*(4), 619–633.

Smith, N. J., & Levy, R. (2013). The effect of word predictability on reading time is logarithmic. *Cognition*, *128*(3), 302–319.

Snell, J., van Leipsig, S., Grainger, J., & Meeter, M. (2018). OB1-reader: A model of word recognition and eye movements in text reading. *Psychological Review*, *125*(6), 969–984.

Staub, A. (2015). The Effect of Lexical Predictability on Eye Movements in Reading: Critical Review and Theoretical Interpretation. *Language and Linguistics Compass*, *9*(8), 311–327.

Stuellein, N., Radach, R. R., Jacobs, A. M., & Hofmann, M. J. (2016). No one way ticket from orthography to semantics in recognition memory: N400 and P200 effects of associations. *Brain Research*, *1639*, 88–98.

Taylor, W. L. (1953). "Cloze" procedure: A new tool for measuring readability. *Journalism Quarterly*, *30*(4), 415–433.

Traxler, M. J., Foss, D. J., Seely, R. E., Kaup, B., & Morris, R. K. (2000). Priming in Sentence Processing : Intralexical Spreading Activation , Schemas , and Situation Models. *Journal of Psycholinguistic Research*, *29*(6), 581–595.

Wang, H., Pomplun, M., Chen, M., & Ko, H. (2010). Estimating the effect of word predictability on eye movements in Chinese reading using latent semantic analysis and transitional probability. *Quarterly Journal of Experimental Psychology*, *63*(7), 37–41.

# Appendix A

Table A. Mean values (*SD* in parentheses) and *F*-scores of the controlled variables

|  | HH | HL | LH | LL | F |
|---|---|---|---|---|---|
| AS verb-adjective | 0.00 (0.00) | 0.02 (0.13) | 0.00 (0.00) | 0.02 (0.14) | 0.68 |
| CA verb-adjective | 29.50 (10.44) | 25.93 (18.54) | 28.28 (14.73) | 27.40 (22.66) | 0.31 |
| Closed-class word 1 | | | | | |
| Length | 3.25 (1.26) | 3.38 (1.43) | 3.28 (1.22) | 3.38 (1.39) | 0.10 |
| Frequency | 3.10 (2.41) | 3.65 (2.82) | 2.98 (2.25) | 3.08 (2.39) | 0.61 |
| Closed-class word 2 | | | | | |
| Length | 3.38 (1.15) | 3.40 (1.15) | 3.30 (1.04) | 3.38 (1.08) | 0.06 |
| Frequency | 2.68 (1.59) | 2.55 (1.41) | 2.58 (1.41) | 2.25 (1.46) | 0.62 |
| Closed-class word 3 | | | | | |
| Length | 3.35 (0.89) | 3.43 (0.75) | 3.53 (1.09) | 3.40 (1.13) | 0.23 |
| Frequency | 2.63 (1.90) | 2.50 (1.54) | 2.88 (1.64) | 2.63 (1.55) | 0.36 |

*Note*. AS = direct associative strength; CA = number of common associates; HH = high number of CA between both primes and noun; HL = High number of CA between verb and noun and low number of CA between adjective and noun; LH = Low number of CA between verb and noun and high number of CA between adjective and noun: LL = low number of CA between both primes and target.